\def\year{2023}\relax
\documentclass[letterpaper]{article} 
\usepackage{aaai23}  
\usepackage{times}  
\usepackage{helvet} 
\usepackage[hyphens]{url}  
\usepackage{graphicx} 
\usepackage{amsmath}
\usepackage{amssymb}
\usepackage{natbib}  
\usepackage{caption} 

\usepackage{booktabs}
\usepackage{multirow}
\usepackage{algorithm}
\usepackage{xcolor}
\usepackage{tabularx}

\title{Heuristic Search For Physics-Based Problems: Angry Birds in PDDL+}
\author{
    Wiktor Piotrowski\textsuperscript{\rm 1},
    Yoni Sher\textsuperscript{\rm 1},
    Sachin Grover\textsuperscript{\rm 1},
    Roni Stern\textsuperscript{\rm 1,2},
    Shiwali Mohan\textsuperscript{\rm 1}
}
\affiliations{
    \textsuperscript{\rm 1} Palo Alto Research Center, CA, USA\\
    \textsuperscript{\rm 2} Ben-Gurion University of the Negev, Beer-Sheva, Israel\\
    \{wiktorpi,sgrover,smohan\}@parc.com,sternron@bgu.ac.il,yoni.sandsher@gmail.com\\
}
\begin{document}

\maketitle

\begin{abstract}

This paper studies how a domain-independent planner and combinatorial search can be employed to play Angry Birds, a well established AI challenge problem. To model the game, we use PDDL+, a planning language for mixed discrete/continuous domains that supports durative processes and exogenous events. The paper describes the model and identifies key design decisions that reduce the problem complexity. In addition, we propose several domain-specific enhancements including heuristics and a search technique similar to preferred operators. Together, they alleviate the complexity of combinatorial search. We evaluate our approach by comparing its performance with dedicated domain-specific solvers on a range of Angry Birds levels. The results show that our performance is on par with these domain-specific approaches in most levels, even without using our domain-specific search enhancements.

%
\end{abstract}

\section{Introduction}


Angry Birds, a wildly popular mobile game, is an open challenge problem for AI \cite{renz2019ai} that requires reasoning about sequential actions in a continuous world with discrete exogenous events. Different versions of the game have proven to be NP-Hard, PSPACE-complete, and EXPTIME-hard~\cite{stephenson2020computational}, and the reigning world champion is still a human. 
The game's simple layout and mechanics teamed with human players' innate spatio-temporal reasoning and forward state prediction makes for a challenging task. However, where humans excel, AI agents struggle. Small errors in initial assumptions can result in scores and states drastically different from the player's predictions. Angry Birds requires a holistic understanding of each individual level and its relevant characteristics. Understanding the relevance of all features and the sum of all of its parts is not a common trait of AI approaches. Angry Birds is a difficult and fascinating challenge for autonomous agents, solving it would prove a significant milestone in AI.

In this work, we present the first successful game playing agent for Angry Birds that uses a domain-independent planner and combinatorial search.
Most existing planning languages, such a STRIPS \cite{fikes1971strips}, PDDL \cite{mcdermott1998pddl}, and PDDL2.1 \cite{fox2003pddl2}, lack features such as exogenous activity to capture the Angry Birds games. In this paper, we study how the game mechanics can be encoded using PDDL+~\cite{fox2006modelling}, a rich planning language that is designed for mixed discrete/continuous domains. 
PDDL+ enables modeling the physics-based dynamics such as effect of gravity, collision between objects, and trajectory of the bird using equations of motion and their computational approximations (e.g., Bhaskara's sine approximation formula).


Our first contribution is a model of Angry Birds using PDDL+, which includes several important modeling choices that ensure solution in reasonable time. We employ time discretization techniques to reason about continuous aspects of the domain which presents the challenge of combinatorial search. The second contribution of this work is the development of several search enhancements designed specifically to constrain solution space in the domain. These enhancements include domain-specific heuristics and a ``preferred states'' mechanism similar to the preferred operators technique~\cite{richter2009preferred}. Third contribution is through a comprehensive evaluation against some of the state-of-the-art agents for Angry Birds ~\cite{borovicka2014datalab,wang2017description}.
Our evaluation demonstrates that our techniques can solve a greater diversity of Angry Bird levels compared to other agents thus, showcasing that proposed domain-independent search strategies and domain-specific heuristics make search more efficient.
Please note, this is a preliminary exploratory work incorporating the initial modeling and searching techniques for discrete-continuous domains, such as basic features of birds of point and shoot. Since, submitting this work we have expanded the capabilities to incorporate some of the special features, such as, second tap to initialize special power of birds, that will be part of the forthcoming submissions.

The paper is organized as follows. We begin by providing relevant background on the Angry Birds game and PDDL+. This is followed by an analysis of different planning formalisms and their limitations for this domain. Next, we present the design of our PDDL+ model for Angry Birds. Then, we describe how this PDDL+ model is used within a game playing agent, and propose several domain-specific search enhancements. 
Finally, we close with an experimental evaluation of our approach on a variety of Science Birds problems comparing our results with state-of-the-art baseline agents from previous competitions.
For the convenience of the readers we have provided the complete domain, a problem file and the solution generated as part of the supplementary material with the submission.

\section{Background}
\begin{figure}
\begin{center}
\includegraphics[trim = 0 100 0 150, clip, width=0.4\textwidth]{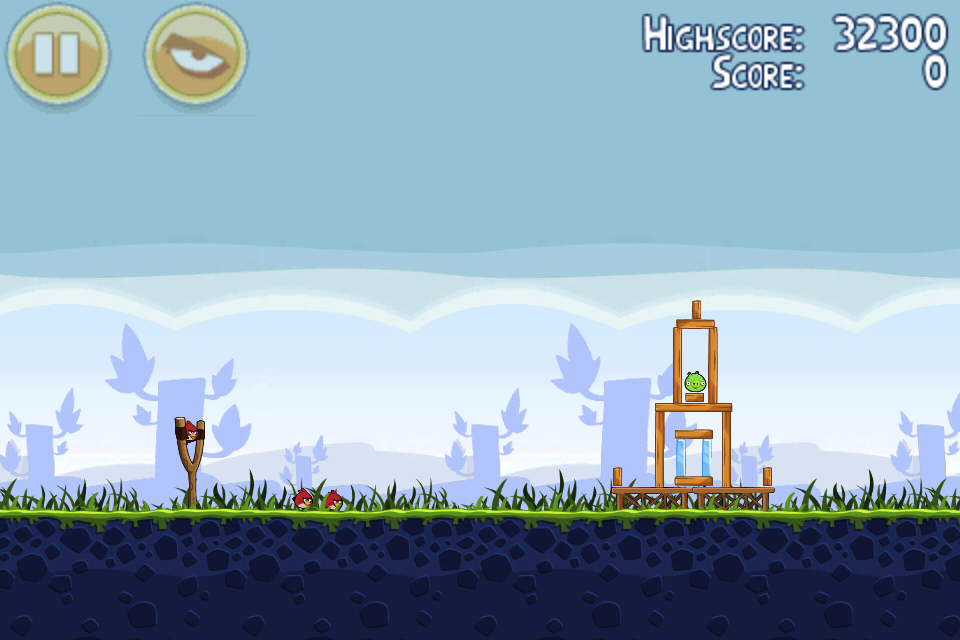}
\end{center}
\caption{Trivial Angry Birds level.}
\label{fig:easy-level}
\end{figure}
\begin{figure}
\begin{center}
\includegraphics[trim = 0 20 0 20, clip, width=0.4\textwidth]{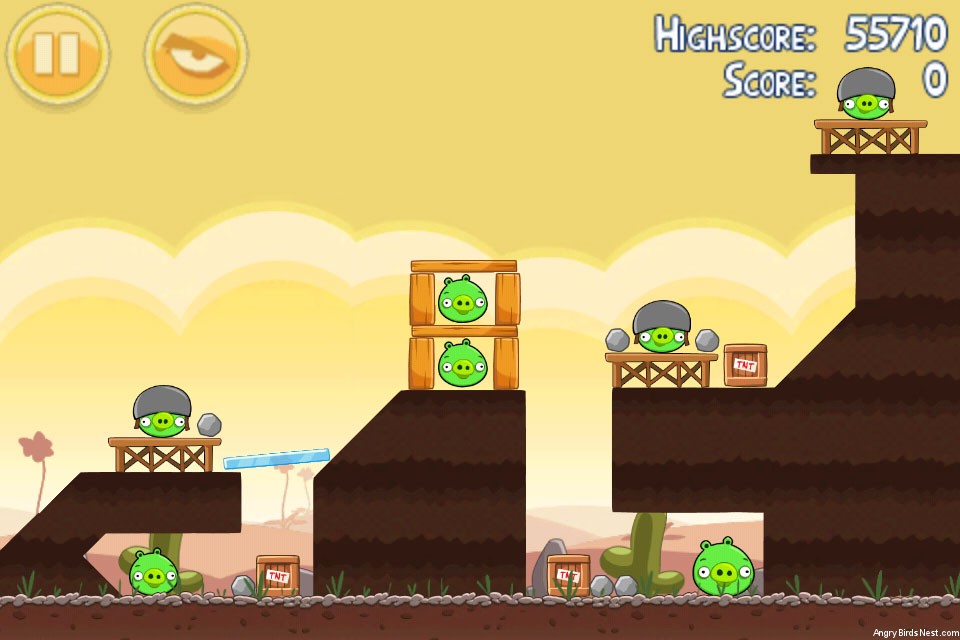}
\end{center}
\caption{Difficult Angry Birds level.}
\label{fig:hard-level}
\end{figure}

\setlength{\belowcaptionskip}{-3pt}

PDDL+~\cite{fox2006modelling} is an extension of the well-known Planning Domain Description Language (PDDL)~\cite{mcdermott1998pddl}, that maps the planning constructs to a Hybrid Automata~\cite{henzinger2000theory} for improved expressiveness with numeric fluents and metrics, instantaneous actions, exogenous {\em events} and durative changes called {\em processes}.
PDDL+ is designed to model and plan for problems requiring both discrete mode switches, continuous flows with exogenous environmental changes.
It builds on the expressiveness of its predecessors modeling languages, encapsulating the entire set of features from PDDL2.1~\cite{fox2003pddl2} and adding timed-initial literals from PDDL2.2~\cite{edelkamp2004pddl2}.

In PDDL+ the domain $\mathcal{D}$ for the game environment is a tuple $\langle F, R, A, E, P \rangle$ where
$F$ is a set of discrete state variables;
$R$ is a set of numeric state variables;
$A$ is a set of actions the agent can execute;
$E$ is a set of instantaneous events triggered in the environment directly or indirectly due to actions of the agent;
$P$ is a set of durative process that may be active in the environment or triggered by the agent.
A state of the environment $s$ is a complete assignment of values to the state variables $F \cup R$. Please note, that the domain of numeric state variables does not include $\bot$ value, i.e., none of the variables are undefined in the state as the environment and the transitions are completely deterministic.
A planning problem $\Pi = \langle \mathcal{D}, s_I, G \rangle$, where $s_I$ is the initial state, and $s_I \in S$ the set of possible states in the environment; $G$ is the goal condition for the environment. 


\subsubsection{The Angry Birds Game} consists of several different levels. Every level has some allocated numbers of birds that are launched using a slingshot in an attempt to hit pigs sitting inside structures built using {\em platforms} and {\em blocks}.
Pigs can be killed by a direct hit from a bird, or indirectly by falling blocks, explosions, or falling from a height.
Every level consists of different arrangement of structures and a number of pigs to be killed.
Platforms are indestructible floating objects, while blocks can be destroyed or toppled over. Blocks come in various shapes, sizes, and materials. 
For example, blocks made of ice break easily, stone blocks are harder to break, and TNT crates (shown in fig.~\ref{fig:hard-level}) explodes on impact, destroying close objects and launching others in the air.
Some birds have special abilities, activated by tapping on the screen during flight. 
While these special abilities are needed to pass some levels, we do not model them in this paper for simplicity. 


Aim of the game is to kill the pigs using minimum number of birds and maximize the game score. A well-known strategy of angry birds is to hit the weakest point of the structure, to maximize the impact on the pigs and kill them quickly. The composition of each level can drastically affect the choice of strategy for playing the game. 
For example, Figure \ref{fig:easy-level} shows a simple level with red birds and a single pig protected by a weak structure. 
To pass this level, it is sufficient to shoot a bird directly towards the pig, causing the structure to collapse on it. 
In contrast, Figure \ref{fig:hard-level} shows a more difficult level where the only viable strategy is executing precise shots that set off an explosive chain reaction. 
Thus, an intelligent agent playing Angry Birds needs to be able to predict future states of the world that extend beyond the immediate and direct consequences of the agent's actions. 
As the game is solved in discrete and continuous state space, it has become the game of choice for a long running yearly competition, AI Birds~\cite{renz2015aibirds}, organized by Australian National University at the IJCAI conference.

\section{Modeling Angry Birds}
To create a domain for Angry Birds, we need to model the flight of the bird, collisions between structures, explosions, and structure collapse after collisions and explosions. In this section we look at these aspects in some detail. First, we begin with a brief discussion on why we chose PDDL+ as our modeling language, then describe the PDDL+ components used to model different objects in the game, and finally the PDDL+ components used to model the game dynamics. 

\begin{table}[h]
\centering
\small
\begin{tabular}{c|c|c|c|}
\cline{2-4}
 & \textbf{PDDL+} & \textbf{PDDL2.1} & \textbf{PDDL2.2} \\ \hline
\multicolumn{1}{|c|}{Bird Flight}              & $\checkmark$ & $\checkmark$ &  \\ \hline
\multicolumn{1}{|c|}{Collisions}               & $\checkmark$ &   & $\checkmark$   \\ \hline
\multicolumn{1}{|c|}{Explosions}               & $\checkmark$ &   & $\checkmark$   \\ \hline
\multicolumn{1}{|c|}{Structure collapse}       & $\checkmark$ &   & $\checkmark$   \\ \hline
\end{tabular}%
\caption{Language support for crucial Angry Birds features. \textbf{FSTRIPS+} similar to \textbf{PDDL2.1} is also able to model only the Bird Flight.}
\label{tab:lang-comparison}
\end{table}

\begin{table*}[!ht]
    \centering
    \begin{tabular}{c|l|c|l}
                                & Predicate & Domain & Comments \\ \toprule
    \multirow{5}{*}{Birds}      & \texttt{bird\_type} & $\mathbb{Z}$ & Different types of birds such as 0 is for the Red bird \\
                                    & \texttt{x\_bird, y\_bird} & $\mathbb{R}$ & XY location of the bird \\
                                    & \texttt{v\_bird, vx\_bird, vy\_bird} & $\mathbb{R}$ & Cumulative and XY component of velocity of of the bird \\
                                    & \texttt{m\_bird} & $\mathbb{R}^+$ & Mass of the bird. \\
                                    & \texttt{bird\_id} & $\mathbb{Z}$ & Sequence in which the birds will be fired. \\ \midrule
    \multirow{2}{*}{Pigs}      & \texttt{x\_pig, y\_pig} & $\mathbb{R}$ & XY location of the pig \\
                                    & \texttt{r\_pig, m\_pig} & $\mathbb{R}^+$ & Radius \& Mass of the pig \\ \midrule
    \multirow{3}{*}{Blocks}    & \texttt{x\_block, y\_block} & $\mathbb{R}$ & XY location of the block \\
                                    & \texttt{block\_height, block\_width} & $\mathbb{R}^+$ & Height and Width of the block \\
                                    & \texttt{block\_mass} & $\mathbb{R}^+$ & Mass of the block \\ \midrule
    \multirow{3}{*}{Platforms} & \texttt{x\_platform, y\_platform} & $\mathbb{R}$ & XY location of the platform \\
                                    & \texttt{platform\_width} & $\mathbb{R}^+$ & Width of the platform.\\
                                    & \texttt{platform\_height} & $\mathbb{R}^+$ & Height of the platform.\\
    \bottomrule
    \end{tabular}
    \caption{Modeling different objects in Angry Birds.}
    \label{tab:model_predicates}
\end{table*}

Different aspects of Angry Birds can be modelled with different levels of expressiveness. For example, PDDL2.1 \cite{fox2003pddl2} can be used to model the flight of the bird using numeric fluents similar to an FSTRIPS+ \cite{ramirez2017numerical}. Table \ref{tab:lang-comparison} provides a comparison of PDDL2.1, PDDL2.2 \cite{edelkamp2004pddl2}, FSTRIPS+, and PDDL+, and what parts of the game can be modelled using them. PDDL2.2 provides a close competition for modeling, however, PDDL+ provides support for all parts of the game.

PDDL+ planning problems are difficult to solve due to immense search spaces, and complex system dynamics. Indeed, planning in hybrid domains is challenging because, apart from the state space explosion caused by discrete state variables, the continuous variables cause the reachability problem to become undecidable~\cite{alur95algorithmic}. 
Thus, clever design choices when modeling in PDDL+ are often crucial to solving PDDL+ problems, arguably more so than for any other planning domain definition languages. 

\subsection{Angry Birds Objects in PDDL+}


The objects in the Angry Birds domain are separated into four types: \textit{birds, pigs, blocks, and platforms}. 
Note that we did not list the slingshot, which is where the birds are launched from, as an object in the domain. 
Instead of explicitly modeling the slingshot, we assign the coordinates of the at-rest slingshot to every bird that is about to be launched.
Avoiding modeling unnecessary objects helps in maintaining a light-weight domain definition. 
Table \ref{tab:model_predicates} describes different predicates that were used to model birds, pigs, blocks and platforms. The state is defined using the XY location of each object and their specific properties, e.g., mass or radius.

\subsection{Angry Birds Dynamics in PDDL+}

The PDDL+ domain of Angry Birds features a variety of dynamics which dictate the change and evolution of the system. This, however, creates planning problems with vast search spaces and large branching factors, making them computationally difficult to solve. Thus, mitigating the issues of solvability and efficiency was at the forefront of the domain's development at each step of the process, and is evident in the resulting PDDL+ model's composition.

\subsubsection{Launching a Bird}
Our PDDL+ model of Angry Birds contains a single action -- launching a bird from the slingshot at a chosen angle. 
Launching of the bird is split into three separate phases: selecting the launch velocity, finding the launch angle, and releasing the bird. 
In our model, we fix the launch velocity to its maximum possible value to decrease the solution search space. This decision is motivated by the fact that maximum velocity shots provide widest range of targets and the final impact velocity of the bird is directly proportional to damage it causes to the objects. Moreover, the same level of reachability can be modeled by changing the angle of slingshot alone instead of modulating the speed and angle together.


Next, we address the construction of the action responsible for selecting the launch angle. In a naive approach, one could encode a set of instantaneous actions that increase or decrease the angle, supplemented by an action releasing the bird from the slingshot.
In this approach, the search space will include many cycles, corresponding to increasing and decreasing the launch angle by equal amounts. 
Instead, we pair the release action with a supporting process that continuously adjusts the angle as soon as a new bird is placed on the slingshot, ready to be launched. 
This follows the Theory of Waiting~\cite{mcdermott2003reasoning}, which sees the agent idly waiting until the world evolves into a favorable state in which to execute actions. 
This encoding reduces the number of decision points, cycles, and branching factor, mitigating state space explosion.

Concretely, the model includes an \emph{increase\_angle} process for increasing the launch angle and a \emph{release\_bird} action (fig.~\ref{fig:action-pa-twang}) whose effects assign values to the vertical and horizontal velocity variables based on the angle. The velocities are then used to model the ballistic flight of the birds. 
Once a bird is launched, another process is triggered. This \textit{flying} process (fig.~\ref{fig:process-flying}) models ballistic flight of the active bird, updating its velocity and location over time, according to the governing equations of motion. 
Note that this process represents the domain's non-linear behavior, which is a major obstacle in PDDL+ planning as many planners struggle to handle complex dynamics.



\begin{figure}
\begin{center}
    \fontsize{8pt}{10pt}\selectfont
\begin{verbatim}
 (:action release_bird
    :parameters (?b - bird)
    :precondition (and
        (= (active_bird) (bird_id ?b))
        (not (angle_adjusted))
        (not (bird_released ?b)) )
    :effect (and
        (assign (vy_bird ?b) 
            (* (v_bird ?b) SIN(angle)))
        (assign (vx_bird ?b) 
            (* (v_bird ?b) COS(angle)))
        (bird_released ?b) (angle_adjusted)) )
\end{verbatim}
\caption{PDDL+ action for launching a bird. Since trigonometric functions are not included in PDDL+, model relies on Bhaskara's small-angle approximations for SIN and COS.}
\label{fig:action-pa-twang}
\end{center}
\end{figure}

\subsubsection{Events}
All the objects interact with the environment and other objects through events. Below, we discuss three different classes of events for modeling collisions, motion of the blocks, and auxiliary events at the start and end of each shot. These events model complex behaviors of destruction, explosion and structure collapse when hit by birds. In our planning model, all the post launch system dynamics are caused by events interacting with the effects of the flying process.

\begin{figure}
\begin{center}
    \fontsize{8pt}{10pt}\selectfont
\begin{verbatim}
 (:process flying
    :parameters (?b - bird)
    :precondition (and
        (bird_released ?b)
        (= (active_bird) (bird_id ?b))
        (> (y_bird ?b) 0) )
    :effect (and
        (decrease (vy_bird ?b) 
            (* #t (gravity) ))
        (increase (y_bird ?b) 
            (* #t (vy_bird ?b)))
        (increase (x_bird ?b) 
            (* #t (vx_bird ?b)))) )
\end{verbatim}
\caption{PDDL+ process of a bird's flight.}
\label{fig:process-flying}
\end{center}
\end{figure}

Collisions are the first class of events that are highlighted in an Angry Birds problem. The main aim of the game is to kill pigs by directly colliding with them, causing nearby explosions, or destroying blocks such that pigs are killed by collapsing structures or falling from heights. Each of those cases is a crucial tactic in Angry Birds, and any accurate model of the game must encompass such mechanics. 

\begin{figure*}
\begin{center}
\includegraphics[width=0.23\textwidth]{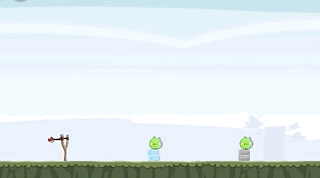}
\includegraphics[width=0.23\textwidth]{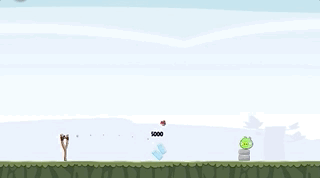}
\includegraphics[width=0.23\textwidth]{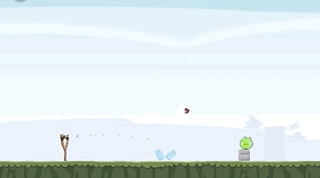}
\includegraphics[width=0.23\textwidth]{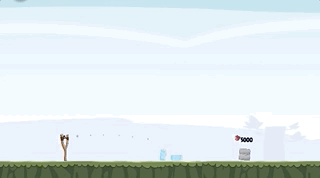}
\end{center}
\caption{Two pigs-one bird plan in execution.}
\label{fig:double-hit}
\end{figure*}

As an example event, direct interactions between a bird and a pig are modeled based on elastic sphere collisions in two dimensions. Pigs are very fragile and disappear upon almost any contact from another object. Thus, after a collision, only the bird's resulting velocity needs to be calculated, according to the angle-free elastic collision formula:
\begin{equation} \label{eq:collision}
\vec{v}_{b}' = \vec{v}_{b} - \frac{2m_{p}}{m_{b}+m_{p}} \frac{\langle \vec{v}_{b} - \vec{v}_{p},\vec{x}_{b} - \vec{x}_{p}\rangle}{||\vec{x}_{b} - \vec{x}_{p}||^2}(\vec{x}_{b} - \vec{x}_{p})
\end{equation}
where $\vec{v}_b$ and $\vec{v}_p$ are the velocity of the bird and pig, respectively, $\vec{x}_b$ and $\vec{x}_p$ are their respective positions, $m_b$ and $m_p$ are their respective masses, and the angle brackets denote an inner product of two vectors. 
Translating this equation into a PDDL+ event allows us to model behavior beyond the obvious solutions (i.e., aiming directly at an exposed pig). It enables the planner to find highly innovative solutions such as deliberately and directly killing two pigs with one bird (fig.~\ref{fig:double-hit}).
Similarly, we model events for collisions with the grounds, which also enables finding trick shots in which the bird bounces off the ground to hit a pig or a block.

\begin{figure*}
\begin{center}
\includegraphics[width=0.23\textwidth]{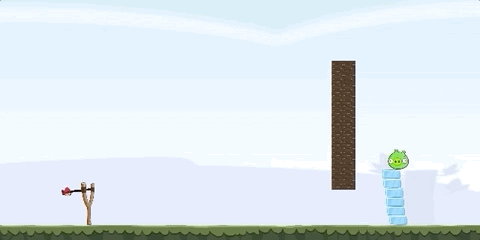}
\includegraphics[width=0.23\textwidth]{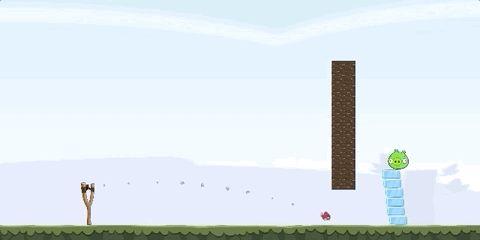}
\includegraphics[width=0.23\textwidth]{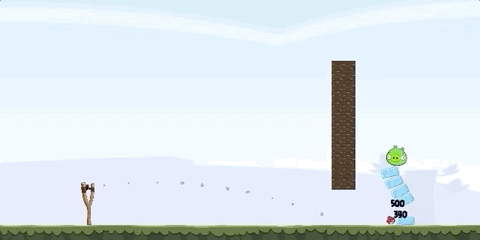}
\includegraphics[width=0.23\textwidth]{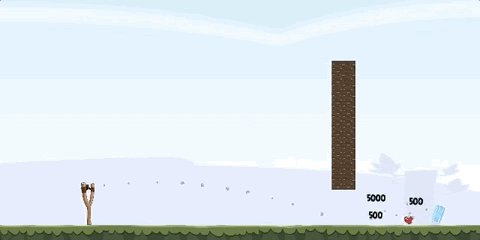}
\end{center}
\caption{Solution requiring bouncing off the ground and collapsing a structure to kill the target pig.}
\label{fig:bounce-hit}
\end{figure*}

The ground in Angry Birds is often overlooked when playing the game but it can prove useful when aiming at difficult targets. As such, an event (seen in fig~\ref{fig:ground-event} is defined in the domain that influences the bird's trajectory after bouncing off the ground, based on energy lost during the collision (i.e., ground damper). Figure~\ref{fig:bounce-hit} shows how a ground-bounce event helps in killing a pig when a direct hit is impossible.

\begin{figure}[!ht]
\begin{center}
\begingroup
    \fontsize{8pt}{10pt}\selectfont
\begin{verbatim}
 (:event collision_ground
  :parameters (?b - bird)
  :precondition (and
     (= (active_bird) (bird_id ?b))
     (<= (y_bird ?b) 0)  )
  :effect (and
     (assign (y_bird ?b) 1)
     (assign (vy_bird ?b) 
     (* (* (vy_bird ?b) -1)(ground_damper)))
     (assign (bounce_count ?b) 
     (+ (bounce_count ?b) 1))) )
 \end{verbatim}
\endgroup
 \caption{Event modeling a bird bouncing off the ground.}
\label{fig:ground-event}
 \end{center}
\end{figure}

\smallskip
Most interactions in a level occurs between birds and blocks in the scene. Blocks form structures which act as a protection barrier for targeted pigs. Thus, modeling collisions with blocks is vital to playing Angry Birds. However, similarly to pig collisions, we opt to only define the event's effects in terms of the impact it has on the bird's velocity and the blocks stability and life values. There are two events, modeling bird-block collisions with the distinction based on whether the block is stable enough to withstand the impact without being knocked out of place or destroyed entirely. In such cases, the bird bounces back off the block, otherwise it continues moving forward with a diminished velocity. 

Modeling the motion of blocks after impact and chains of block-block collisions would be prohibitively difficult, requiring a process for tracking each individual block's change in position and rotation over time, as well as a set of events to account for secondary interactions. 
These additions would only marginally improve the domain's accuracy, though the planner would experience a drastic drop in performance, having to keep track of dozens of simultaneous non-linear processes in every state after a collision. Instead, we model one central block-block interaction, namely dislodging or destroying blocks supporting larger structures. In such cases, an event is triggered for every supported block, reducing its stability value to 0, repositioning the block to the ground, and adjusting its life value to account for fall damage. Analogously, we defined another event for killing pigs which sit atop collapsing block structures. 

The two final block-related events concern exploding TNT crates. Any impact upon the crate causes an explosion destroying all objects in its immediate vicinity and displacing objects further away. In our PDDL+ model, we incorporate two simple events modeling the explosive destruction of pigs and blocks near a TNT crate. As previously indicated, we omit movement of objects caused by explosions. 


Platforms are indestructible objects which the agent needs to reason with to avoid poor results. However, almost no meaningful interactions occur between birds and platforms, and thus our domain contains an event modeling this type of collision but only to discourage the agent from shooting at platforms. The event deters the agent by nullifying the bird's velocity and expiring it without any impact on the level.

Finally, we encode a supporting event which models the end of life for an active bird. In the game, a bird expires when it moves beyond the limits of the scene or when it stops moving. Modeling the exact timing of the end of a bird's life is quite difficult. Instead, we define the birds usefulness in terms of the number of bounces off other objects. Each bounce partially reduces the bird's momentum, and its collision impact becomes negligible after three bounces. Therefore, we define a termination event for the active bird once it has bounced three times. The result of this event is that the next bird is placed on the slingshot for launch.

\textbf{Goals}
While the goal of Angry Birds is to destroy all the pigs, we consider a simpler formulation for individual planning problems. The simplified problem splits the original scenario into single-bird episodes with the goal of killing at least one pig. During evaluations we start with this simplified approach and use a direct trajectory calculator as a contingency strategy when the planning phase fails to find a solution.
Table \ref{table:domain_stats} gives a breakdown of the number of events, processes and actions in a domain, and the number of objects in a typical level. As noted above, Angry Birds is an atypical domain as the agent only has one action (release bird). 

\begin{table}
    \centering
    \small
    \begin{tabular}{l|r|rrr}
    \toprule
        Level & Objects & Actions & Events & Processes \\
        \midrule
        22 & 7-11   & 5(1)  & 19-114(17)        & 3(3) \\
        25 & 8-12   & 5(1)  & 20-88(17)         & 3(3) \\
        36 & 8-12   & 5(1)  & 30-114(17)        & 3(3) \\
        45 & 9-13   & 5(1)  & 37-162(17)        & 3(3) \\
        46 & 11-15  & 9(1)  & 59-199(17)        & 5(3) \\
        53 & 10-14  & 5(1)  & 28-163(17)        & 3(3) \\
        54 & 8-12   & 5(1)  & 30-114(17)        & 3(3) \\
        57 & 11-15  & 5(1)  & 65-220(17)        & 3(3) \\
        55 & 29-138 & 13(1) & 430-13772(17)     & 7(3) \\
         \bottomrule
    \end{tabular}
    \caption{Number of objects and happenings in the grounded domain (and in the domain template). The main computational load is checking whether events have occurred. }
    \label{table:domain_stats}

\end{table}

\section{Solving Angry Birds with a PDDL+ Planner}
In this section, we describe how we used the PDDL+ modeling of Angry Birds specified above to design an Angry Birds playing agent called \emph{Hydra}. We detail how Hydra interacts with the game, the PDDL+ planner it uses, and search techniques we implemented to improve its efficiency. 



\subsection{From Angry Birds to PDDL+}

The game playing API represents an Angry Birds level as a list of labeled objects and their locations. Hydra supplements this information with background knowledge of the game and its objects, such as the mass of the different birds. Then, it automatically translates all the collective relevant information of the current level into a PDDL+ problem file. 

Next, it uses a PDDL+ planner to generate a plan that kills at least one pig using one bird. If no plan is found in 30s, we execute a default non-planning action, namely a direct shot at a random pig. 
If a plan has been found, Hydra performs the plan. After performing an action, Hydra observes the updated state of the game, again via the game playing API. Then, it generates a corresponding PDDL+ problem as before, and generates a new plan from the current state. This continues until the level ends, either passing the level or failing to do so before exhausting all the available birds. 

\subsubsection{PDDL+ Planners}
Any state-of-the-art PDDL+ planners can be used for solving the Angry Birds game, such as, SMTPlan+~\cite{cashmore2016compilation}, UPMurphi~\cite{della2009upmurphi}, DiNo~\cite{piotrowski2016heuristic} and ENHSP~\cite{scala2016interval}.
However, during experimentation we faced several challenges due to costly domain-independent heuristics and lack of ability to incorporate other heuristics for improved results in the game.
Thus, we used our own domain-independent PDDL+ planner in Hydra, which is written in Python and has been used successfully incorporated for several other domains.
Now we will discuss some of the search algorithms that are supported by our planner and the two heuristics that we evaluated for the domain.

\subsection{Search and Heuristics}
Our domain-independent PDDL+ planner supports several search algorithms including depth-first search (DFS), breadth-first search (BFS), and Greedy Best-First Search (GBFS)\footnote{Our planner also support A*~\cite{hart1968formal}, but since we do not aim to find optimal solutions, there is no benefit in using A*. Nevertheless, we performed an initial evaluation of using A* in Hydra, but the results were subpar.}.
GBFS requires heuristic function to evaluate explored states. We propose two heuristics for our domain. 

\subsubsection*{The Score Heuristic} While the agent aims to \emph{solve} levels in the game, i.e., kill all pigs, the game itself associates a score to each state. Specifically, destroying pigs and blocks increases the score of the current level. The score heuristic, denoted $H_S$, uses exactly this score as a heuristic.
The agent needs to analyze the opportune angles at which to release the slingshot and evaluate distant effects. The heuristic is calculated based on the number of pigs and blocks the bird 
will hit multiplied by a constant value.




\subsubsection*{The Proximity Heuristic} The second heuristic we used calculates the active bird's proximity to the nearest pig, as a function of flight time and direction. This heuristic, denoted $H_P$, is computed as follows. 
\begin{eqnarray}
    \vec{d} = \vec{x}_{target} - \vec{x}_{bird} && \text{vector from bird to pig}   \\
    \hat{d} = \frac{\vec{d}}{\lVert \vec{d} \rVert} & &\text{unit vector of same}   \\
    H_P(s) = \frac{\rVert \vec{d} \rVert}{\vec{v}_{bird} \cdot \hat{d}} && \text{flight time to pig} 
\end{eqnarray}
Where $ \vec{x}_{target}, \vec{x}_{bird} $ and $ \vec{v}_{bird} $ are the coordinate vectors of the pig, the bird and the velocity of the bird respectively, $ \lVert \vec{x} \rVert $ is the Euclidean norm, and $ \cdot $ is the inner product. 
Using this heuristic results in giving preference to states where the bird is not only closer to a pig, but also heading towards it, reducing the effort spent on distant trajectories. As a consequence, this heuristic also gives preference to low trajectories.
The heuristic is admissible as the straight-line distance to the target underestimating the flight time. 


\subsection{Helpful States}
Orthogonal to the choice of heuristics, our system incorporates domain-specific strategies into the search process.
This mechanism, that we refer to as \emph{helpful states}, is based on the ``helpful operators'' mechanism from classical planning~\cite{richter2009preferred}. 
Helpful operators in classical planning is a mechanism that prioritizes the use of a subset of the operators while planning. 
Porting this mechanism to our case requires some adaptation, since the number of actions in our domain is very limited.

Our \emph{helpful states} mechanism accepts some definition of what a preferred state is, and then prioritize expanding such states. In our implementation we defined a preferred state as one in which the active bird is on a trajectory that is expected to hit a pig or a TNT block. Checking if the active bird is on a trajectory to hit an object is done by using the equation for ballistic motion to create a function $ y_t = f_i(x_t) $ for each potential target $i$ (pig or TNT block). Given a state, we can calculate whether the bird's $y$ is close to the desired value, and if so, mark it as a preferred state.
The search strategy can also be understood as doing Monte-Carlo rollout \cite{chaslot2008progressive} only on a smaller set of states, instead of performing a complete rollout on all the search states.

Throughout the search we maintain two open lists, one that only includes preferred states, and one that includes all states. During the search, we alternate between expanding states from the open list with the preferred states and the regular open list. 
To ensure that preferred states are reached early, we also mark all states where the bird has not yet been launched as preferred. This ensures the planner reasons with the full range of possible launch angles. 
Note that this ``preferred states'' mechanism is fairly general, and allows different definitions of preferred states, and multiple open lists with different priorities. This is a topic for future research.

\section{Experiments}

\begin{table*}[t]
    \begin{center}
    \begin{tabular}{p{0.29\textwidth}p{0.29\textwidth}p{0.29\textwidth}}
         \includegraphics[width=0.29\textwidth]{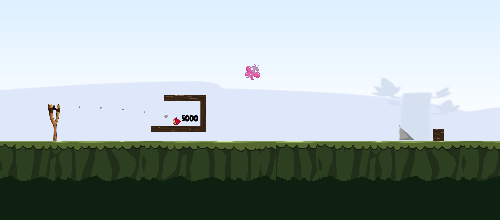} & \includegraphics[width=0.29\textwidth]{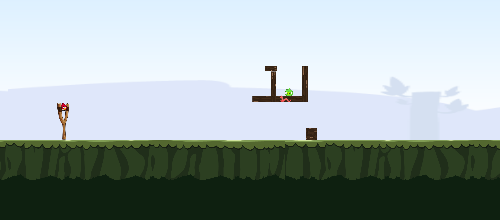} &
         \includegraphics[width=0.29\textwidth]{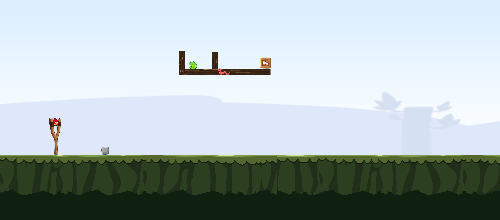} \\
          22: A level requiring a single low-trajectory shot. & 25: A level requiring a single high-trajectory shot. & 36: In this level, the TNT is too far away to kill the pig.  \\
         \includegraphics[width=0.29\textwidth]{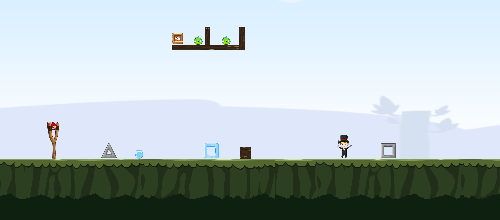} & 
         \includegraphics[width=0.29\textwidth]{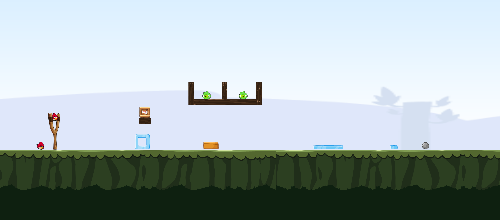} & 
         \includegraphics[width=0.29\textwidth]{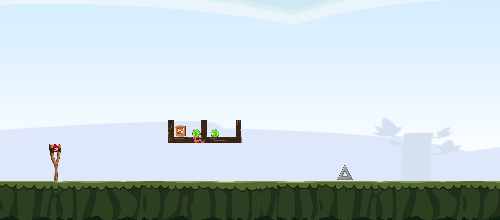} \\
          45: To kill both pigs with one bird, must hit TNT.  & 46: the TNT is a decoy, must hit each pig with a bird.  & 53: Must hit TNT to win, but it is a difficult shot.  \\
          \includegraphics[width=0.29\textwidth]{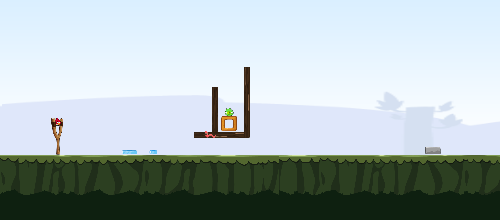} & 
          \includegraphics[width=0.29\textwidth]{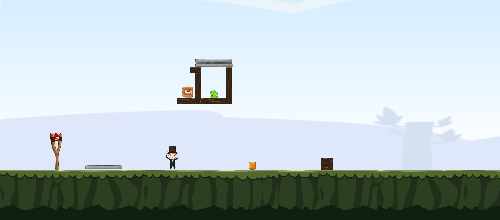} & 
          \includegraphics[width=0.29\textwidth]{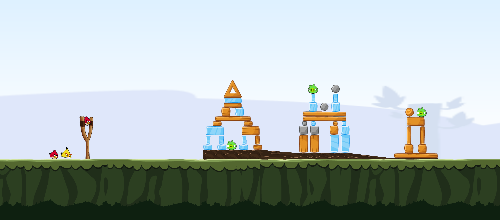} \\
           54: Bird must bounce off far platform wall to hit pig.  & 57: TNT provides a solution to an `impenetrable' box.  & 55: A complex level. 
    \end{tabular}
    \captionof{figure}{A screenshot example of each type of level layout}
    \label{fig:levels}
    \end{center}
\end{table*}

We conducted a set of experiments to evaluate the efficacy of various planning-related design decisions explored in this paper to develop Hydra. Particularly, in experiment 1, we studied how different search perform when playing Angry Birds using the PDDL+ model introduced in this paper. Next, in experiment 2, we measured the impact of different heuristics have on controlling the search.  These experiments were conducting on a benchmark set of Angry Birds problems. We also present results from three other domain-specific Angry Birds agents on the benchmark set as a way to situate Hydra's performance. 

\subsubsection{Benchmark Levels}
Our benchmark set of problems contains \emph{simple levels} that contain a single bird and a relatively small number of other objects, and \emph{complex levels} that contain multiple birds and more than 50 objects. The evaluation is designed to test the agents in situations requiring varied strategies. The simple levels require high accuracy shots 
while the complex levels require a high level of physics reasoning about interaction with blocks, but can often be passed by simple curated rules.

In total, we experimented with 9 types of levels, which are numbered 22, 25, 36, 45, 46, 53, 54, 57, and 55. 
Every evaluated agent attempted to solve 25 randomly generated levels of each type. 
An example layout for each type of level can be found in figure \ref{fig:levels}. 
Note that the first 8 level types are simple levels and the last one is a complex level.

\subsubsection{Competing Agents}
We selected a baseline agent by ANU and two former champions of the IJCAI competition. 
\begin{itemize}
    \item \textbf{ANU Baseline}~\cite{stephenson2017creating} is a baseline agent developed by the IJCAI AI Birds competition organizers. This baseline agent targets a randomly selected pig with each active bird, and aims to shoot at the selected pig while disregarding all other objects in the scene.
    \item \textbf{DataLab}~\cite{borovicka2014datalab} is the 2014 \& 2015 champion of the IJCAI Angry Birds competition. DataLab's selects from 4 predefined strategies (destroy pigs, target TNT, target round blocks, destroy structures), based on predicted utilities.     
    \item \textbf{Eagle's Wing}~\cite{wang2017description} is the 2016-2018 champion of the ICJAI Angry Birds competition. Eagle's Wing selects from 5 predefined strategies (pig-shooter, TNT, most blocks, high round objects, and bottom building blocks), based on predicted utility. Eagle's Wing supplements these strategies with a trained XGBoost~\cite{chen2016xgboost} model to optimize its performance.
    \item \textbf{Hydra}: We evaluated different variations of Hydra that use different search strategies, breadth-first search (Hy.,BFS), depth-first search (Hy.,DFS), and greedy-best first search (Hy.,GBFS). Three variations of Hy.,GBFS were tested each leveraging different heuristics discussed in this paper: score heuristic ($H_S$), proximity heuristic ($H_P$), and helpful states ($HS$).  
    
\end{itemize}
All agents are designed to work with the ANU Science Birds framework~\cite{renz2015aibirds}.



\subsubsection{Results 1}
\begin{table}
\small
\centering
\begin{tabular}{@{ }l>{\centering\arraybackslash}p{0.03\columnwidth}>{\centering\arraybackslash}p{0.03\columnwidth}>{\centering\arraybackslash}p{0.03\columnwidth}>{\centering\arraybackslash}p{0.03\columnwidth}>{\centering\arraybackslash}p{0.03\columnwidth}>{\centering\arraybackslash}p{0.03\columnwidth}>{\centering\arraybackslash}p{0.03\columnwidth}>{\centering\arraybackslash}p{0.03\columnwidth}>{\centering\arraybackslash}p{0.03\columnwidth}@{ }}
\toprule
                    & \multicolumn{9}{c}{Levels}                 \\ 
Agent               & 22 & 25 & 36 & 45 & 46 & 53 & 54 & 57 & 55 \\ \midrule
Hy., BFS            & \textbf{21} & \textbf{21} & 18 & \textbf{25} & 18 & 15 & 11 & \textbf{25} & 4  \\
Hy., DFS            & \textbf{21} & 16 & 16 & 21 & 10 & 13 & 11 & 17 & 2  \\
Hy., GBFS($H_S$)    & \textbf{21} & \textbf{21} & 20 & \textbf{25} & 14 & 15 & 14 & \textbf{25} & 1  \\
Hy., GBFS($H_P$)    & \textbf{21} & \textbf{21} & 24 & \textbf{25} & \textbf{25} & \textbf{25} & \textbf{16} & \textbf{25} & 4  \\
Hy., GBFS($HS$)     & \textbf{21} & \textbf{21} & \textbf{25} & \textbf{25} & 24 & \textbf{25} & \textbf{16} & \textbf{25} & 9  \\
\midrule
ANU Baseline        & 18 & 0  & 0  & 16 & 0  & 2  & 0  & 11 & \textbf{17} \\
Datalab             & 15 & 8  & 0  & 24 & 3  & 0  & 0  & 22 & 16 \\
Eaglewings          & 2  & 6  & 0  & 23 & 2  & 0  & 1  & 21 & 15 \\ \bottomrule
\end{tabular}
\caption{Evaluation results for each agent, reporting on number of levels passed.} 
\label{tab:solved-results}
\end{table}

First, we measure the number of levels \emph{passed} by each agent. This metric ensure that we study not only that the agents can plan but are able to find a plan that can be executed in the environment for successful goal achievement. Table~\ref{tab:solved-results} shows the number of levels passed from each of the level templates in our benchmark, by each agent.

The results show several clear trends. First, all Hydra agents outperformed the other agents in all the simple levels. In some cases, e.g., level type 36, all GBFS variants of HYDRA passed at least 20 out of 25 levels of this type while none of the non-Hydra agents were able to pass any levels. On the other hand, the baseline agents outperformed all Hydra agents in the complex levels. 

Hydra's heuristics guide the planner through the physics of the problem, and are independent of level structure. As a result, Hydra is accurate and versatile, and it can reason with any level composition. Hydra achieves best performance in simple levels requiring highly accurate shots.

In contrast, non-Hydra agents were developed for playing human-designed levels, which are readily solved by a small number of fixed strategies focused on destroying complex structures. Low-accuracy strategies fare well on densely-populated levels (i.e., level 55 in the evaluation). Agents developed for levels with hand-crafted choices of strategy, and requiring only low-accuracy shots, do not perform well when the underlying assumptions for those strategies are broken (i.e., randomly generated level compositions). 


\begin{table}[tbh]
\resizebox{1\columnwidth}{!}{
\begin{tabular}{@{}lrrrrrrrrr@{}}
\toprule
Level & 22    & 25    & 36  & 45  & 46  & 53  & 54    & 57  & 55 \\
\midrule
Nodes/s & 1,610 & 1,295 & 930 & 624 & 732 & 747 & 1,202 & 634 & 36 \\
\bottomrule
\end{tabular}
}
\caption{Rate of node expansion, in nodes per second, by Hydra using GBFS with helpful states (GBFS (HS)), for each of level type in our benchmark (top row).} 
\label{tab:nodes-per-sec}
\end{table}
Hydra does not perform well in complex levels since the sheer number of objects present require higher computational resources to reason about them. However, this suggests that smarter search and heuristics may allow our agent to also solve the complex levels - an extension we are considering for future work. Consider Table~\ref{tab:nodes-per-sec} that lists the number of nodes expanded per second when using Hydra with GBFS and the helpful states technique. As can be seen, node generation in the complex level type (type 55) is much slower, going from 1,610 nodes/s in simple level type 22 to only 36 nodes/s for the complex level type 55. Solving complex levels with domain-independent planning requires stronger heuristics and perhaps other search techniques.

\begin{table}
\centering
\begin{tabular}{@{}rrr|rrr@{}}
\toprule
\multicolumn{1}{c}{}      & \multicolumn{1}{c}{}    & \multicolumn{1}{c}{}    & \multicolumn{3}{c}{GBFS}                                                 \\
\multicolumn{1}{c}{Level} & \multicolumn{1}{c}{BFS} & \multicolumn{1}{c}{DFS} & \multicolumn{1}{c}{$H_S$} & \multicolumn{1}{c}{$H_P$} & \multicolumn{1}{c}{$HS$} \\
\midrule
22                        & 11,468                  & 20,000                  & 11,495                 & 10,081                 & 8,445                  \\
25                        & 36,681                  & 46,108                  & 35,432                 & 21,382                 & 9,668                  \\
36                        & 32,111                  & 42,482                  & 32,753                 & 18,383                 & 1,970                  \\
45                        & 5,257                   & 21,341                  & 5,279                  & 2,264                  & 1,826                  \\
46                        & 31,517                  & 40,871                  & 31,317                 & 15,485                 & 2,674                  \\
53                        & 26,340                  & 36,199                  & 29,366                 & 14,909                 & 2,760                  \\
54                        & 34,564                  & 42,100                  & 35,115                 & 23,578                 & 15,526                 \\
57                        & 6,126                   & 17,974                  & 7,298                  & 2,378                  & 1,753                  \\
55                        & 1,208                   & 1,489                   & 1,454                  & 970                    & 707               \\ \bottomrule    
\end{tabular}
\caption{Average total number of nodes expanded by the different Hydra agents for different level types.}
\vspace{-0.1cm}
\label{tab:expanded}
\end{table}

\subsubsection{Results 2}
Next, we study how various heuristics implemented in Hydra impact plan search. To measure search efficacy, we recorded the number of nodes expanded during planning. Table~\ref{tab:expanded} shows the average number of nodes expanded by each of the Hydra agents we considered. Here we see that all of our implemented heuristics are effective as they reduce the number of nodes expanded during search. 

While intuitively appealing, the score heuristic ($H_S$) turns out to be ineffective since the PDDL+ model of the game has only one action (release the slingshot at an opportune angle), and the action has distant effects. The score is not a direct consequence of the action, but rather it is computed far in the future once the bird collides with other objects.

Helpful states mechanism is the most effective; in some cases it expands an order fewer nodes before finding a solution (see the row corresponding to level 53 for example). Returning to the results in Table \ref{tab:solved-results}, we see that search efficiency significantly impacts Hydra's performance. GBFS with various heuristics can solve more levels that BFS and DFS. Between the different search enhancements, the helpful states technique is clearly beneficial in most cases.

\section{Conclusion}
In this work, we presented a novel approach based on domain-independent planning for Angry Birds, a popular mobile games and a challenging AI testbed~\cite{renz2019ai}. 
To capture the complex dynamics of the system, we modeled the game using PDDL+, a rich planning language developed for hybrid systems supporting exogenous activity in planning domains via discrete events and continuous processes. 
We discuss how expressiveness of PDDL+ can capture complex, physics-based dynamics of Angry Birds. We implemented this PDDL+ model within a game playing planning agent, and explored several domain-specific search enhancements, such as novel heuristics and the ``helpful-states'' search strategy. Such improvements were useful to efficiently solve the challenging Angry Birds domain, despite a PDDL+ model focused on reducing system complexity and only incorporating a single action.
Our evaluation showed promising results, with our agent able to pass more levels than multiple champions of the Angry Birds AI competition, in most types of levels we considered. It highlights that domain-independent planning and combinatorial search is a viable approach for solving this AI benchmark. 

Despite promising results, our current agent can struggle with complex levels composed of several objects on the scene. Additionally, our implementation does not consider the full range of the Angry Birds game, ignoring birds with special abilities and various types of pigs. Future work will address both limitations: develop better heuristics and search algorithms and extending our domain model to include birds' special powers. Future work will also compare Hydra with emerging AI agents such as BamBirds and Agent X\cite{lutalo2022agent}, which improve shot selection by better balancing between exploration and exploitation.

Our work demonstrates that PDDL+ is useful for capturing the dynamics of challenging, real-world domains. 
However, our experience with Angry Birds demonstrates that solving planning problems in rich PDDL+ domains is a challenge for existing planners. The development of solutions for PDDL+ by the community has been slow. One of the reasons for this is that PDDL+ benchmark domains have been largely re-purposed from existing domains, which result in problems that could be addressed by PDDL2.1 techniques. 
This work also serves as a call-to-arms for research into PDDL+ planners and heuristics, both domain-independent and model-specific.

\section{Acknowledgements}

This work was supported by the DARPA SAIL-ON program under contract HR001120C0040. The
views and conclusions in this document are those of the authors and should not be interpreted as
representing the official policies, either expressly or implied, of the Defense Advanced Research
Projects Agency or the U.S. Government.

\bibliography{library}
\end{document}